\documentclass[11pt,a4paper]{article}
\usepackage[hyperref]{emnlp2020}
\usepackage{times}
\usepackage{latexsym}

\usepackage{amsmath,amssymb}
\usepackage{amsfonts}
\usepackage{multirow,multicol}
\usepackage{pifont}
\usepackage{graphicx}

\usepackage{microtype}

\aclfinalcopy 

\title{Vector Projection Network for Few-shot Slot Tagging in Natural Language Understanding}

\author{Su Zhu, Ruisheng Cao, Lu Chen \and Kai Yu\thanks{\ \ The corresponding author is Kai Yu.}\\
  MoE Key Lab of Artificial Intelligence\\
  SpeechLab, Department of Computer Science and Engineering\\
  Shanghai Jiao Tong University, Shanghai, China\\
  {\tt \{paul2204,211314,chenlusz,kai.yu\}@sjtu.edu.cn} \\}

\date{}

\begin{document}
\maketitle
\begin{abstract}

Few-shot slot tagging becomes appealing for rapid domain transfer and adaptation, motivated by the tremendous development of conversational dialogue systems. In this paper, we propose a vector projection network for few-shot slot tagging, which exploits projections of contextual word embeddings on each target label vector as word-label similarities. Essentially, this approach is equivalent to a normalized linear model with an adaptive bias. The contrastive experiment demonstrates that our proposed vector projection based similarity metric can significantly surpass other variants. Specifically, in the five-shot setting on benchmarks SNIPS and NER, our method outperforms the strongest few-shot learning baseline by $6.30$ and $13.79$ points on F$_1$ score, respectively. Our code will be released at \url{https://github.com/sz128/few_shot_slot_tagging_and_NER}.
\end{abstract}

\section{Introduction}

Natural language understanding (NLU) is a key component of conversational dialogue systems, converting user's utterances into the corresponding semantic representations~\cite{wang2005spoken} for certain narrow domain (e.g., \emph{booking hotel}, \emph{searching flight}). As a core task in NLU, slot tagging is usually formulated as a sequence labeling problem~\cite{mesnil2015using,sarikaya2016overview,liu2016attention}.

Recently, motivated by commercial applications like Amazon Alexa, Apple Siri, Google Assistant, and Microsoft Cortana, great interest has been attached to rapid domain transfer and adaptation with only a few samples~\cite{Bapna2017towards}. Few-shot learning approaches~\cite{fei2006one,vinyals2016matching} become appealing in this scenario ~\cite{fritzler2019few,geng2019induction,hou2020few}, where a general model is learned from existing domains and transferred to new domains rapidly with merely few examples (e.g., in one-shot learning, only one example for each new class).

The similarity-based few-shot learning methods have been widely analyzed on classification problems~\cite{vinyals2016matching,snell2017prototypical,sung2018learning,yan2018few,yu2018diverse,sun2019hierarchical,geng2019induction,yoon2019tapnet}, which classify an item according to its similarity with the representation of each class. These methods learn a domain-general encoder to extract feature vectors for items in existing domains, and utilize the same encoder to obtain the representation of each new class from very few labeled samples (\emph{support set}). This 
scenario has been successfully adopted in the slot tagging task by considering both the word-label similarity and temporal dependency of target labels~\cite{hou2020few}. Nonetheless, it is still a challenge to devise appropriate word-label similarity metrics for generalization capability.

In this work, a vector projection network is proposed for the few-shot slot tagging task in NLU. To eliminate the impact of unrelated label vectors but with large norm, we exploit projections of contextual word embeddings on each normalized label vector as the word-label similarity. Moreover, the half norm of each label vector is utilized as a threshold, which can help reduce false positive errors. 


One-shot and five-shot experiments on slot tagging and named entity recognition (NER)~\cite{hou2020few} tasks show that our method can outperform various few-shot learning baselines, enhance existing advanced methods like TapNet~\cite{yoon2019tapnet,hou2020few} and prototypical network~\cite{snell2017prototypical,fritzler2019few}, and achieve state-of-the-art performances.

Our contributions are summarized as follows:
\begin{itemize}
    \item We propose a vector projection network for the few-shot slot tagging task that utilizes projections of contextual word embeddings on each normalized label vector as the word-label similarity.
    \item We conduct extensive experiments to compare our method with different similarity metrics (e.g., dot product, cosine similarity, squared Euclidean distance). Experimental results demonstrate that our method can significantly outperform the others.
\end{itemize}

\section{Related Work}

One prominent methodology for few-shot learning in image classification field mainly focuses on metric learning~\cite{vinyals2016matching,snell2017prototypical,sung2018learning,oreshkin2018tadam,yoon2019tapnet}. The metric learning based methods aim to learn an effective distance metric. It can be much simpler and more efficient than other meta-learning algorithms~\cite{munkhdalai2017meta,mishra2018a,finn2017model}.

As for few-shot learning in natural language processing community, researchers pay more attention to classification tasks, such as text classification~\cite{yan2018few,yu2018diverse,sun2019hierarchical,geng2019induction}. Recently, few-shot learning for slot tagging task becomes popular and appealing. \citet{fritzler2019few} explored few-shot NER with the prototypical network. \citet{hou2020few} exploited the TapNet and label dependency transferring for both slot tagging and NER tasks. Compared to these methods, our model can achieve better performance in new domains by utilizing vector projections as word-label similarities.

\section{Problem Formulation}

\begin{figure}[]
\centering
\includegraphics[width=0.48\textwidth]{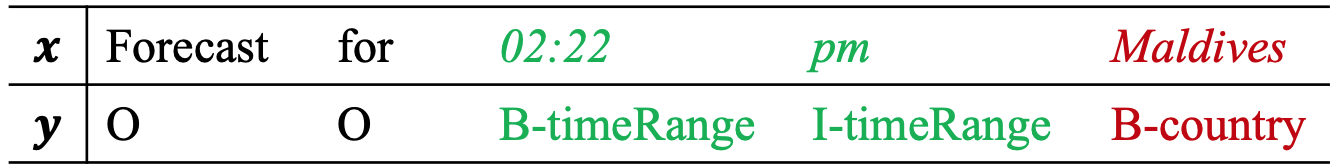}
\caption{A data sample in domain \texttt{GetWeather}.}
\label{fig:data_sample}
\end{figure}

We denote each sentence $\boldsymbol{x} = (x_1, \cdots, x_{|\boldsymbol{x}|})$ as a word sequence, and define its label sequence as $\boldsymbol{y} = (y_1, \cdots, y_{|\boldsymbol{x}|})$. An example for slot tagging in domain \texttt{GetWeather} is provided in Fig \ref{fig:data_sample}. For each domain $\mathcal{D}$, it includes a set of $(\boldsymbol{x}, \boldsymbol{y})$ pairs, i.e., $\mathcal{D}=\{(\boldsymbol{x}^{(i)}, \boldsymbol{y}^{(i)})\}_{i=1}^{|\mathcal{D}|}$.

In the few-shot scenario, the slot tagging model is trained on several source domains $\{\mathcal{D}_1, \mathcal{D}_2, \cdots, \mathcal{D}_M\}$, and then directly evaluated on an unseen target domain $\mathcal{D}_t$ which only contains few labeled samples~(\emph{support set}). The support set, $S=\{(\boldsymbol{x}^{(i)}, \boldsymbol{y}^{(i)})\}_{i=1}^{|S|}$, usually includes $k$ examples (K-shot) for each of N labels (N-way). Thus, the few-shot slot tagging task is to find the best label sequence $\boldsymbol{y}^*$ given an input query $\boldsymbol{x}$ in target domain $\mathcal{D}_t$ and its corresponding support set $S$, 
\begin{equation}
    \boldsymbol{y}^* = \arg \max _{\boldsymbol{y}} p_{\theta}(\boldsymbol{y}|\boldsymbol{x}, S)
\end{equation}
where $\theta$ refers to parameters of the slot tagging model, the $(\boldsymbol{x}, \boldsymbol{y})$ pair and the support set are from the target domain, i.e., $(\boldsymbol{x}, \boldsymbol{y}) \sim \mathcal{D}_t$ and $S \sim \mathcal{D}_t$.

The few-shot slot tagging model is trained on the source domains to minimise the error in predicting labels conditioned on the support set,
\begin{equation*}
    \theta = \arg \max _{\theta} \sum_{m=1}^M \sum_{(\boldsymbol{x}, \boldsymbol{y}) \sim \mathcal{D}_m,\ S \sim \mathcal{D}_m} \log p_{\theta}(\boldsymbol{y}|\boldsymbol{x}, S)
\end{equation*}


\section{Vector Projection Network}

In this section, we will introduce our model for the few-shot slot tagging task.

\subsection{Few-shot CRF Framework}

Linear Conditional Random Field (CRF)~\cite{sutton2012introduction} considers the correlations between labels in neighborhoods and jointly decodes the most likely label sequence given the input sentence~\cite{yao2014recurrent,ma-hovy-2016-end}. The posterior probability of label sequence $\mathbf{y}$ is computed via:
\begin{align*}
\psi_\theta(\boldsymbol{y}, \boldsymbol{x}, S) &= \sum_{i=1}^{|\boldsymbol{x}|} (f_T(y_{i-1}, y_i) + f_E(y_i, \boldsymbol{x}, S))\\
p_{\theta}(\boldsymbol{y}|\boldsymbol{x}, S) &= \frac{
\text{exp}(\psi_\theta(\boldsymbol{y}, \boldsymbol{x}, S))}{\sum_{\boldsymbol{y}'}\text{exp}(\psi_\theta(\boldsymbol{y}', \boldsymbol{x}, S))
}
\end{align*}
where $f_T(y_{i-1}, y_i)$ is the transition score and $f_E(y_i, \boldsymbol{x}, S)$ is the emission score at the $i$-th step. 

The transition score captures temporal dependencies of labels in consecutive time steps, which is a learnable scalar for each label pair. To share the underlying factors of transition between different domains, we adopt the Collapsed Dependency Transfer (CDT) mechanism~\cite{hou2020few}.

The emission scorer independently assigns each word a score with respect to each label $y_i$, which is defined as a word-label similarity function:
\begin{equation}
    f_E(y_i, \boldsymbol{x}, S) = \textsc{Sim}(E(\boldsymbol{x})_i, \mathbf{c}_{y_i})
\end{equation}
where $E$ is a contextual word embedding function, e.g., BLSTM~\cite{graves2012supervised}, Transformer~\cite{vaswani2017attention}, $\mathbf{c}_{y_i}$ is the label embedding of $y_i$ which is extracted from the support set $S$. In this paper, we adopt a pre-trained BERT model~\cite{devlin2019bert} as $E$.

Various models are proposed to extract label embedding $\mathbf{c}_{y_i}$ from $S$, such as matching network~\cite{vinyals2016matching}, prototypical network~\cite{snell2017prototypical} and TapNet~\cite{yoon2019tapnet}. Take the prototypical network as an example, each prototype~(label embedding) is defined as the mean vector of the embedded supporting points belonging to it:
{
\begin{equation}
    \mathbf{c}_{y_i} = \frac{1}{N_{y_i}}\sum_{j=1}^{|S|}\sum_{k=1}^{|\boldsymbol{x}^{(j)}|} \mathbb{I}\{y_k^{(j)}=y_i\} E(\boldsymbol{x}^{(j)})_k
\end{equation}}where $N_{y_i}=\sum_{j=1}^{|S|}\sum_{k=1}^{|\boldsymbol{x}^{(j)}|} \mathbb{I}\{y_k^{(j)}=y_i\}$ is the number of words labeled with $y_i$ in the support set.

\subsection{Vector Projection Similarity}

For the word-label similarity function, we propose to exploit vector projections of word embeddings $\mathbf{x}_i$ on each normalized label vector $\mathbf{c}_{k}$:
{
\begin{equation}
    \textsc{Sim}(\mathbf{x}_i, \mathbf{c}_{k}) = \mathbf{x}_i^{\top}\frac{\mathbf{c}_{k}}{||\mathbf{c}_{k}||}
\end{equation}}Different with the dot product used by~\citet{hou2020few}, it can help eliminate the impact of $\mathbf{c}_{k}$'s norm to avoid the circumstance where the norm of $\mathbf{c}_{k}$ is enough large to dominate the similarity metric. In order to reduce false positive errors, the half norm of each label vector is utilized as an adaptive bias term:
{
\begin{equation}
    \textsc{Sim}(\mathbf{x}_i, \mathbf{c}_{k}) = \mathbf{x}_i^{\top}\frac{\mathbf{c}_{k}}{||\mathbf{c}_{k}||} - \frac{1}{2}||\mathbf{c}_{k}||
\end{equation}}

\subsection{Explained as a Normalized Linear Model}

A simple interpretation for the above vector projection network is to learn a distinct linear classifier for each label. We can rewrite the above formulas as a linear model:
{
\begin{equation}
    \textsc{Sim}(\mathbf{x}_i, \mathbf{c}_{k}) = \mathbf{x}_i^{\top}\mathbf{w}_k + b_k
    \label{eqn:norm_linear}
\end{equation}}where $\mathbf{w}_k=\frac{\mathbf{c}_{k}}{||\mathbf{c}_{k}||}$ and $b_k=-\frac{1}{2}||\mathbf{c}_{k}||$. The weights are normalized as $||\mathbf{w}_k||=1$ to improve the generalization capability of the few-shot model. Experimental results indicate that vector projection is an effective choice compared to dot product, cosine similarity, squared Euclidean distance, etc.

\section{Experiment}

We evaluate the proposed method following the data split~\footnote{\url{https://atmahou.github.io/attachments/ACL2020data.zip}} provided by \citet{hou2020few} on SNIPS~\cite{coucke2018snips} and NER datasets. It is in the \emph{episode} data setting~\cite{vinyals2016matching}, where each episode contains a support set (1-shot or 5-shot) and a batch of labeled samples. For slot tagging, the SNIPS dataset consists of 7 domains with different label sets: Weather (We), Music (Mu), PlayList (Pl), Book (Bo), Search Screen (Se), Restaurant (Re) and Creative Work (Cr). For NER, 4 different datasets are utilized to act as different domains: CoNLL-2003 (News)~\cite{sang2003introduction}, GUM (Wiki)~\cite{zeldes2017gum}, WNUT-2017 (Social)~\cite{derczynski2017results} and OntoNotes (Mixed)~\cite{pradhan2013towards}. More details of the data split are shown in Appendix \ref{app:sec:data_statistic}.

For each dataset, we follow \citet{hou2020few} to select one target domain for evaluation, one domain for validation, and utilize the rest domains as source domains for training. We also report the average F$_1$ score at the episode level. For each experiment, we run it ten times with different random seeds. The training details are illustrated in Appendix \ref{app:sec:training}.

 \begin{table*}[htbp]
    \small
    \centering
\begin{tabular}{c|ccccccccc}
    \hline
    & \textbf{Model} & \textbf{We} & \textbf{Mu} & \textbf{Pl} & \textbf{Bo} & \textbf{Se} & \textbf{Re} & \textbf{Cr} & \textbf{Avg.} \\ 
    \hline
    \multirow{7}{*}{\textbf{1-shot}} & SimBERT & 36.10 & 37.08 & 35.11 & 68.09 & 41.61 & 42.82 & 23.91 & 40.67 \\
    & TransferBERT & 55.82 & 38.01 & 45.65 & 31.63 & 21.96 & 41.79 & 38.53 & 39.06 \\
    & L-WPZ(ProtoNet)+CDT+PWE & 71.23	& 47.38	& 59.57	& 81.98	& 69.83	& 66.52	& 62.84	& 65.62 \\
    & L-TapNet+CDT+PWE & 71.53	& {60.56}	& 66.27	& \textbf{84.54}	& 76.27	& 70.79	& 62.89	& 70.41 \\
    \cline{2-10}
    & L-TapNet+CDT+VP (ours) & 71.65 & \textbf{61.73} & 63.97 & 83.34 & 74.00 & 71.91 & \textbf{71.02} & 71.09  \\
    & ProtoNet+CDT+VP (ours) & \textbf{73.56} & 58.40 & {68.93} & 82.32 & \textbf{79.69} & {73.40} & {70.25} & \textbf{72.37} \\
    & L-ProtoNet+CDT+VP (ours) & 73.19 & 58.62 & 68.26 & 83.54 & 77.88 & \textbf{73.48} & 69.54 & 72.07 \\
    & ProtoNet+CDT+VPB (ours) & 72.65 & 57.35 & 68.72 & 81.92 & 74.68 & 72.48 & 70.04 & 71.12 \\
    & L-ProtoNet+CDT+VPB (ours) & 73.12 & 57.86 & \textbf{69.01} & 82.49 & 75.11 & 73.34 & {70.46} & 71.63 \\
    \hline\hline
    \multirow{7}{*}{\textbf{5-shot}} & SimBERT & 53.46 & 54.13 & 42.81 & 75.54 & 57.10 & 55.30 & 32.38 & 52.96 \\
     & TransferBERT & 59.41 & 42.00 & 46.07 & 20.74 & 28.20 & 67.75 & 58.61 & 46.11 \\
     & L-WPZ(ProtoNet)+CDT+PWE & 74.68 & 56.73 & 52.20 & 78.79 & 80.61 & 69.59 & 67.46 & 68.58 \\
     & L-TapNet+CDT+PWE & 71.64 & 67.16 & 75.88 & 84.38 & 82.58 & 70.05 & 73.41 & 75.01 \\
    \cline{2-10}
     & L-TapNet+CDT+VP (ours) & 78.25 & {67.79} & 70.66 & 86.17 & 75.80 & 78.51 & 75.93 & 76.16  \\
     & ProtoNet+CDT+VP (ours) & {79.88} & {67.77} & {78.08} & {87.68} & \textbf{86.59} & {79.95} & {75.61} & {79.37} \\
     & L-ProtoNet+CDT+VP (ours) & 80.26 & 67.81 & 74.62 & 88.16 & 85.79 & 80.41 & 73.84 & 78.70 \\
     & ProtoNet+CDT+VPB (ours) & {82.91} & {69.23} & {80.85} & {90.69} & 86.38 & {81.20} & \textbf{76.75} & {81.14} \\
     & L-ProtoNet+CDT+VPB (ours) & \textbf{82.93} & \textbf{69.62} & \textbf{80.86} & \textbf{91.19} & 86.58 & \textbf{81.97} & 76.02 & \textbf{81.31} \\
    \hline
\end{tabular}
    \caption{F$_1$ scores on few-shot slot tagging of SNIPS. Results with standard deviations is shown in Appendix \ref{app:sec:res_std}.} 
    \label{tab:1shot_5shot_snips}
\end{table*}
 \begin{table*}[]
    \small
    \centering
\begin{tabular}{ccccccccccc}
    \hline
    \multirow{2}{*}{\textbf{Model}} & \multicolumn{4}{c}{\textbf{1-shot}} &  & \multicolumn{4}{c}{\textbf{5-shot}} & \\
    \cline{2-5} \cline{7-10}
     & \textbf{News} & \textbf{Wiki} & \textbf{Social} & \textbf{Mixed} & \textbf{Avg.} & \textbf{News} & \textbf{Wiki} & \textbf{Social} & \textbf{Mixed} & \textbf{Avg.} \\
    \hline
    SimBERT & 19.22 & 6.91 & 5.18 & 13.99 & 11.32 & 32.01 & 10.63 & 8.20 & 21.14 & 18.00 \\
    TransferBERT & 4.75 & 0.57 & 2.71 & 3.46 & 2.87 & 15.36 & 3.62 & 11.08 & {35.49} & 16.39 \\
    L-TapNet+CDT+PWE & 44.30 & \textbf{12.04} & 20.80 & 15.17 & 23.08 & 45.35 & 11.65 & 23.30 & 20.95 & 25.31 \\
    \hline
    L-TapNet+CDT+VP (ours) & 44.73 & 8.91  & \textbf{30.61} & 29.39 & {28.41} & 50.43 & 8.41  & {29.93} & {37.59} & 31.59  \\
    ProtoNet+CDT+VP (ours) & {44.82} & 11.32 & {26.96} & {29.91} & {28.25} & {54.82} & {16.30} & {27.43} & 33.38 & {32.98} \\
    L-ProtoNet+CDT+VP (ours) & \textbf{45.93}& 8.76  & 29.21 & 32.44 & \textbf{29.09} & 55.68 & 10.39 & 31.39 & 37.83 & 33.82 \\
    ProtoNet+CDT+VPB (ours) & 42.50 & 10.78 & 27.17 & {32.06} & 28.13 & \textbf{57.42} & \textbf{19.48} & {35.06} & {44.45} & \textbf{39.10} \\
    L-ProtoNet+CDT+VPB (ours) & 43.47 & 10.95 & 28.43 & \textbf{33.14} & {29.00} & 56.30 & 18.57 & \textbf{35.42} & \textbf{44.71} & 38.75 \\
    \hline
\end{tabular}
    \caption{F$_1$ scores on few-shot slot tagging of NER. Results with standard deviations is shown in Appendix \ref{app:sec:res_std}.} 
    \label{tab:1shot_5shot_ner}
\end{table*}

 \begin{table}[htbp]
    \small
    \centering
\begin{tabular}{lcccc}
    \hline
    \multirow{2}{*}{$\textsc{Sim}(\mathbf{x}, \mathbf{c})$} & \multicolumn{2}{c}{\textbf{SNIPS}} & \multicolumn{2}{c}{\textbf{NER}} \\
    \cline{2-3} \cline{4-5}
    & \textbf{1-shot} & \textbf{5-shot} & \textbf{1-shot} & \textbf{5-shot} \\
    \hline
    $\mathbf{x}^\top \frac{\mathbf{c}}{||\mathbf{c}||}$ & \textbf{72.37} & 79.37 & \textbf{28.25} & 32.98 \\
    $\mathbf{x}^\top \frac{\mathbf{c}}{||\mathbf{c}||} - \frac{1}{2}||\mathbf{c}||$ &  71.12 & \textbf{81.14} & 28.13 & \textbf{39.10} \\
    \hline
    $\mathbf{x}^\top \mathbf{c}$ & 57.92 & 65.03 & 17.10 & 19.91 \\
    $\frac{\mathbf{x}^\top}{||\mathbf{x}||} \mathbf{c}$ & 63.87 & 71.16 & 16.72 & 23.65 \\
    $\frac{\mathbf{x}^\top}{||\mathbf{x}||} \frac{\mathbf{c}}{||\mathbf{c}||}$ &  34.02 & 39.21 & 10.40 & 12.26 \\
    $\lambda \mathbf{x}^\top \mathbf{c}$ &  48.91 & 68.11 & 5.99 & 21.05 \\
    $-\frac{1}{2}||\mathbf{x}-\mathbf{c}||^2$ &  66.91 & 79.72 & 20.04 & 34.04 \\ 
    \hline
\end{tabular}
    \caption{Comparison among different similarity functions. Results are average F1-scores of all domains.} 
    \label{tab:different_distances}
\end{table}

\subsection{Baselines}

\noindent\textbf{SimBERT:} For each word $x_i$, SimBERT finds the most similar word $x'_k$ in the support set and assign the label of $x'_k$ to $x_i$, according to cosine similarity of word embedding of a fixed BERT.

\noindent\textbf{TransferBERT:} A trainable linear classifier is applied on a shared BERT to predict labels for each domain. Before evaluation, it is fine-tuned on the support set of the target domain.

\noindent\textbf{L-WPZ(ProtoNet)+CDT+PWE:} WPZ is a few-shot sequence labeling model~\cite{fritzler2019few} that regards sequence labeling as classification of each word. It pre-trains a prototypical network~\cite{snell2017prototypical} on source domains, and utilize it to do
word-level classification on target domains without fine-tuning. It is enhanced with BERT, Collapsed Dependency Transfer (CDT) and Pair-Wise Embedding (PWE) mechanisms by~\cite{hou2020few}.

\noindent\textbf{L-TapNet+CDT+PWE:} The previous state-of-the-art method for few-shot slot tagging~\cite{hou2020few}, which incorporates TapNet~\cite{yoon2019tapnet} with BERT, CDT and PWE.

We borrow the results of these baselines from \citet{hou2020few}. ``\textbf{L-}'' means label-enhanced prototypes are applied by using label name embeddings.

\subsection{Main Results}

Table \ref{tab:1shot_5shot_snips} and Table \ref{tab:1shot_5shot_ner} show results on both 1-shot and 5-shot slot tagging of SNIPS and NER datasets respectively. Our method can significantly outperform all baselines including the previous state-of-the-art model. Moreover, the previous state-of-the-art model heavily relies on PWE, which concatenates an input sentence with each sample in the support set and then feeds them into BERT to get pair-wise embeddings. By comparing ``L-TapNet+CDT+PWE'' with ``L-TapNet+CDT+VP'', we can find that our proposed Vector Projection (VP) can achieve better performance as well as higher efficiency. If we incorporate the negative half norm of each label vector as a bias (VPB), F$_1$ score on 5-shot slot tagging is dramatically improved. We 
speculate that 5-shot slot tagging involves multiple support points for each label, thus false positive errors could occur more frequently if there is no threshold when predicting each label. We also find that label name embeddings (``\textbf{L-}') help less in our methods.

\subsection{Analysis}

 \begin{table*}[htbp]
    \small
    \centering
\begin{tabular}{lcccccccccccc}
    \hline
    \multirow{2}{*}{\textbf{Model}} & \multicolumn{3}{c}{\textbf{SNIPS 1-shot}} & \multicolumn{3}{c}{\textbf{SNIPS 5-shot}}& \multicolumn{3}{c}{\textbf{NER 1-shot}} & \multicolumn{3}{c}{\textbf{NER 5-shot}} \\
    \cline{2-4} \cline{5-7} \cline{8-10} \cline{11-13}
    & \textbf{O-X} & \textbf{X-O} & \textbf{X-X} & \textbf{O-X} & \textbf{X-O} & \textbf{X-X} & \textbf{O-X} & \textbf{X-O} & \textbf{X-X} & \textbf{O-X} & \textbf{X-O} & \textbf{X-X} \\
    \hline
     \\
     ProtoNet+CDT & 10815 & 3552 & 17440 & 4802 & 1377 & 6532 & 58498 & 9890 & 35991 & 19344 & 1505 & 9091 \\
     ProtoNet+CDT+VP & 4400 & 3409 & 10638 & 2177 & 1214 & 3610 & 13075 & 29183 & 13893 & 5217 & 6283 & 3595 \\
     ProtoNet+CDT+VPB & 4118 & 3818 & 10959 & 1762 & 1076 & 3343 & 11976 & 26851 & 16032 & 2388 & 6617 & 3280 \\
    \hline
\end{tabular}
    \caption{Error analysis of slot tagging for different error patterns. Numbers are summed over all domains.} 
    \label{tab:error_analysis}
\end{table*}


\noindent \textbf{Ablation Study} For the word-label similarity function $\textsc{Sim}(\mathbf{x}, \mathbf{c})$, we also conduct contrastive experiments between our proposed vector projection and other variants including the dot product ($\mathbf{x}^\top \mathbf{c}$), the projection of label vector on word embedding ($\frac{\mathbf{x}^\top}{||\mathbf{x}||} \mathbf{c}$), cosine similarity ($\frac{\mathbf{x}^\top}{||\mathbf{x}||} \frac{\mathbf{c}}{||\mathbf{c}||}$), squared Euclidean distance ($-\frac{1}{2}||\mathbf{x}-\mathbf{c}||^2$), and even a trainable scaling factor ($\lambda \mathbf{x}^\top \mathbf{c}$)~\cite{oreshkin2018tadam}. 
The results in Table \ref{tab:different_distances} show that our methods can significantly outperform these alternative metrics. We also notice that the squared Euclidean distance can achieve competitive results in the 5-shot setting. Mathematically,
{
\begin{align*}
    -\frac{1}{2}||\mathbf{x}-\mathbf{c}||^2 & =-\frac{1}{2}\mathbf{x}^\top \mathbf{x} + \mathbf{x}^\top \mathbf{c} -\frac{1}{2}\mathbf{c}^\top \mathbf{c}\\ 
    & \approxeq \mathbf{x}^\top \mathbf{c} - \frac{1}{2}\mathbf{c}^\top \mathbf{c}
\end{align*}}where $-\frac{1}{2}\mathbf{x}^\top \mathbf{x}$ is constant with respect to each label and thus omitted. It further consolidates our assumption that $\frac{1}{2}\mathbf{c}^\top \mathbf{c}$ can function as a bias term to alleviate false positive errors.

\begin{figure}[]
\centering
\includegraphics[width=0.45\textwidth]{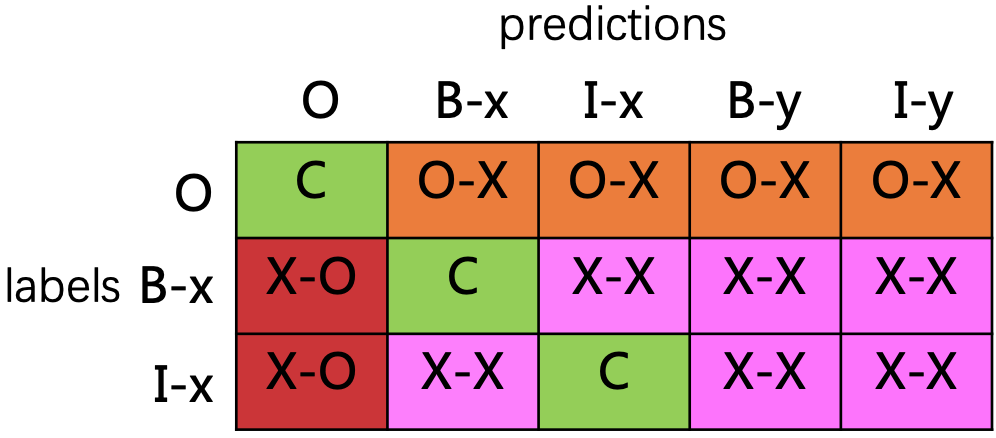}
\caption{Definition of three error types of slot tagging, which are ``O-X'', ``X-O'' and ``X-X''. ``C'' means correct predictions.}
\label{fig:error_types}
\end{figure}

\noindent \textbf{Effect of Vector Projection} We claimed that vector projection could help reduce false positive errors. As illustrated in Figure \ref{fig:error_types}, we classify all wrong predictions of slot tagging into three error types (i.e., ``O-X'', ``X-O'' and ``X-X''), where ``O'' means no slot and ``X'' means a slot tag beginning with `B' or `I'. The error analysis of these three error types are illustrated in Table \ref{tab:error_analysis}. We can find that our methods can significantly reduce wrong predictions of these three types in SNIPS dataset. In NER dataset, our methods can achieve a remarkable reduction in ``O-X'' and ``X-X'', while leading to an increase of ``X-O'' errors. However, the total number of these three errors are reduced by our methods in NER dataset.

\noindent \textbf{Fine-tuning with Support Set} Apart from the few-shot slot tagging focusing on model transfer instead of fine-tuning, we also analyze keeping fine-tuning our models on the support set in Appendix \ref{app:sec:finetune}.

\section{Conclusion}

In this paper, we propose a vector projection network for the few-shot slot tagging task, which can be interpreted as a normalized linear model with an adaptive bias. Experimental results demonstrate that our method can significantly outperform the strongest few-shot learning baseline on SNIPS and NER datasets in both 1-shot and 5-shot settings. Furthermore, our proposed vector projection based similarity metric can remarkably surpass others variants. 

For future work, we would like to add a learnable scale factor for bias in Eqn. \ref{eqn:norm_linear}.

%



\bibliography{emnlp2020}

\begin{thebibliography}{32}
\expandafter\ifx\csname natexlab\endcsname\relax\def\natexlab#1{#1}\fi

\bibitem[{Bapna et~al.(2017)Bapna, T{\"{u}}r, Hakkani{-}T{\"{u}}r, and
  Heck}]{Bapna2017towards}
Ankur Bapna, G{\"{o}}khan T{\"{u}}r, Dilek Hakkani{-}T{\"{u}}r, and Larry~P.
  Heck. 2017.
\newblock Towards zero-shot frame semantic parsing for domain scaling.
\newblock In \emph{INTERSPEECH}, pages 2476--2480.

\bibitem[{{Coucke} et~al.(2018){Coucke}, {Saade}, {Ball}, {Bluche}, {Caulier},
  {Leroy}, {Doumouro}, {Gisselbrecht}, {Caltagirone}, {Lavril}, {Primet}, and
  {Dureau}}]{coucke2018snips}
Alice {Coucke}, Alaa {Saade}, Adrien {Ball}, Th{\'e}odore {Bluche}, Alexandre
  {Caulier}, David {Leroy}, Cl{\'e}ment {Doumouro}, Thibault {Gisselbrecht},
  Francesco {Caltagirone}, Thibaut {Lavril}, Ma{\"e}l {Primet}, and Joseph
  {Dureau}. 2018.
\newblock {Snips Voice Platform: an embedded Spoken Language Understanding
  system for private-by-design voice interfaces}.
\newblock \emph{arXiv preprint arXiv:1805.10190}.

\bibitem[{Derczynski et~al.(2017)Derczynski, Nichols, van Erp, and
  Limsopatham}]{derczynski2017results}
Leon Derczynski, Eric Nichols, Marieke van Erp, and Nut Limsopatham. 2017.
\newblock Results of the {WNUT}2017 shared task on novel and emerging entity
  recognition.
\newblock In \emph{Proceedings of the 3rd Workshop on Noisy User-generated
  Text}, pages 140--147.

\bibitem[{Devlin et~al.(2019)Devlin, Chang, Lee, and
  Toutanova}]{devlin2019bert}
Jacob Devlin, Ming-Wei Chang, Kenton Lee, and Kristina Toutanova. 2019.
\newblock {BERT}: Pre-training of deep bidirectional transformers for language
  understanding.
\newblock In \emph{Proceedings of the 2019 Conference of the North American
  Chapter of the Association for Computational Linguistics: Human Language
  Technologies}, pages 4171--4186.

\bibitem[{Fei-Fei et~al.(2006)Fei-Fei, Fergus, and Perona}]{fei2006one}
Li~Fei-Fei, Rob Fergus, and Pietro Perona. 2006.
\newblock One-shot learning of object categories.
\newblock \emph{IEEE transactions on pattern analysis and machine
  intelligence}, 28(4):594--611.

\bibitem[{Finn et~al.(2017)Finn, Abbeel, and Levine}]{finn2017model}
Chelsea Finn, Pieter Abbeel, and Sergey Levine. 2017.
\newblock Model-agnostic meta-learning for fast adaptation of deep networks.
\newblock In \emph{Proceedings of the 34th International Conference on Machine
  Learning}, pages 1126--1135. JMLR. org.

\bibitem[{Fritzler et~al.(2019)Fritzler, Logacheva, and
  Kretov}]{fritzler2019few}
Alexander Fritzler, Varvara Logacheva, and Maksim Kretov. 2019.
\newblock Few-shot classification in named entity recognition task.
\newblock In \emph{Proceedings of the 34th ACM/SIGAPP Symposium on Applied
  Computing}, pages 993--1000.

\bibitem[{Geng et~al.(2019)Geng, Li, Li, Zhu, Jian, and
  Sun}]{geng2019induction}
Ruiying Geng, Binhua Li, Yongbin Li, Xiaodan Zhu, Ping Jian, and Jian Sun.
  2019.
\newblock Induction networks for few-shot text classification.
\newblock In \emph{Proceedings of the 2019 Conference on Empirical Methods in
  Natural Language Processing and the 9th International Joint Conference on
  Natural Language Processing (EMNLP-IJCNLP)}, pages 3895--3904.

\bibitem[{Graves(2012)}]{graves2012supervised}
Alex Graves. 2012.
\newblock Supervised sequence labelling.
\newblock In \emph{Supervised sequence labelling with recurrent neural
  networks}, pages 5--13. Springer.

\bibitem[{Hou et~al.(2020)Hou, Che, Lai, Zhou, Liu, Liu, and Liu}]{hou2020few}
Yutai Hou, Wanxiang Che, Yongkui Lai, Zhihan Zhou, Yijia Liu, Han Liu, and Ting
  Liu. 2020.
\newblock Few-shot slot tagging with collapsed dependency transfer and
  label-enhanced task-adaptive projection network.
\newblock In \emph{ACL}.

\bibitem[{Kingma and Ba(2014)}]{kingma2014adam}
Diederik~P Kingma and Jimmy Ba. 2014.
\newblock Adam: A method for stochastic optimization.
\newblock \emph{arXiv preprint arXiv:1412.6980}.

\bibitem[{Liu and Lane(2016)}]{liu2016attention}
Bing Liu and Ian Lane. 2016.
\newblock Attention-based recurrent neural network models for joint intent
  detection and slot filling.
\newblock In \emph{17th Annual Conference of the International Speech
  Communication Association}, pages 685--689.

\bibitem[{Ma and Hovy(2016)}]{ma-hovy-2016-end}
Xuezhe Ma and Eduard Hovy. 2016.
\newblock End-to-end sequence labeling via bi-directional {LSTM}-{CNN}s-{CRF}.
\newblock In \emph{the 54th Annual Meeting of the Association for Computational
  Linguistics}, pages 1064--1074.

\bibitem[{Mesnil et~al.(2015)Mesnil, Dauphin, Yao, Bengio, Deng, Hakkani-Tur,
  He, Heck, Tur, Yu et~al.}]{mesnil2015using}
Gr{\'e}goire Mesnil, Yann Dauphin, Kaisheng Yao, Yoshua Bengio, Li~Deng, Dilek
  Hakkani-Tur, Xiaodong He, Larry Heck, Gokhan Tur, Dong Yu, et~al. 2015.
\newblock Using recurrent neural networks for slot filling in spoken language
  understanding.
\newblock \emph{IEEE/ACM Transactions on Audio, Speech, and Language
  Processing}, 23(3):530--539.

\bibitem[{Mishra et~al.(2018)Mishra, Rohaninejad, Chen, and
  Abbeel}]{mishra2018a}
Nikhil Mishra, Mostafa Rohaninejad, Xi~Chen, and Pieter Abbeel. 2018.
\newblock \href {https://openreview.net/forum?id=B1DmUzWAW} {A simple neural
  attentive meta-learner}.
\newblock In \emph{International Conference on Learning Representations}.

\bibitem[{Munkhdalai and Yu(2017)}]{munkhdalai2017meta}
Tsendsuren Munkhdalai and Hong Yu. 2017.
\newblock Meta networks.
\newblock In \emph{Proceedings of the 34th International Conference on Machine
  Learning}, pages 2554--2563. JMLR. org.

\bibitem[{Oreshkin et~al.(2018)Oreshkin, L{\'o}pez, and
  Lacoste}]{oreshkin2018tadam}
Boris Oreshkin, Pau~Rodr{\'\i}guez L{\'o}pez, and Alexandre Lacoste. 2018.
\newblock {TADAM}: Task dependent adaptive metric for improved few-shot
  learning.
\newblock In \emph{Advances in Neural Information Processing Systems}, pages
  721--731.

\bibitem[{Pradhan et~al.(2013)Pradhan, Moschitti, Xue, Ng, Bj{\"o}rkelund,
  Uryupina, Zhang, and Zhong}]{pradhan2013towards}
Sameer Pradhan, Alessandro Moschitti, Nianwen Xue, Hwee~Tou Ng, Anders
  Bj{\"o}rkelund, Olga Uryupina, Yuchen Zhang, and Zhi Zhong. 2013.
\newblock Towards robust linguistic analysis using ontonotes.
\newblock In \emph{Proceedings of the Seventeenth Conference on Computational
  Natural Language Learning}, pages 143--152.

\bibitem[{Sang and De~Meulder(2003)}]{sang2003introduction}
Erik Tjong~Kim Sang and Fien De~Meulder. 2003.
\newblock Introduction to the {CoNLL}-2003 shared task: Language-independent
  named entity recognition.
\newblock In \emph{Proceedings of the Seventh Conference on Natural Language
  Learning at HLT-NAACL 2003}, pages 142--147.

\bibitem[{Sarikaya et~al.(2016)Sarikaya, Crook, Marin, Jeong, Robichaud,
  Celikyilmaz, Kim, Rochette, Khan, Liu et~al.}]{sarikaya2016overview}
Ruhi Sarikaya, Paul~A Crook, Alex Marin, Minwoo Jeong, Jean-Philippe Robichaud,
  Asli Celikyilmaz, Young-Bum Kim, Alexandre Rochette, Omar~Zia Khan, Xiaohu
  Liu, et~al. 2016.
\newblock An overview of end-to-end language understanding and dialog
  management for personal digital assistants.
\newblock In \emph{2016 {IEEE} spoken language technology workshop ({SLT})},
  pages 391--397.

\bibitem[{Snell et~al.(2017)Snell, Swersky, and Zemel}]{snell2017prototypical}
Jake Snell, Kevin Swersky, and Richard Zemel. 2017.
\newblock Prototypical networks for few-shot learning.
\newblock In \emph{Advances in neural information processing systems}, pages
  4077--4087.

\bibitem[{Sun et~al.(2019)Sun, Sun, Zhou, and Lv}]{sun2019hierarchical}
Shengli Sun, Qingfeng Sun, Kevin Zhou, and Tengchao Lv. 2019.
\newblock Hierarchical attention prototypical networks for few-shot text
  classification.
\newblock In \emph{Proceedings of the 2019 Conference on Empirical Methods in
  Natural Language Processing and the 9th International Joint Conference on
  Natural Language Processing (EMNLP-IJCNLP)}, pages 476--485.

\bibitem[{Sung et~al.(2018)Sung, Yang, Zhang, Xiang, Torr, and
  Hospedales}]{sung2018learning}
Flood Sung, Yongxin Yang, Li~Zhang, Tao Xiang, Philip~HS Torr, and Timothy~M
  Hospedales. 2018.
\newblock Learning to compare: Relation network for few-shot learning.
\newblock In \emph{Proceedings of the IEEE Conference on Computer Vision and
  Pattern Recognition}, pages 1199--1208.

\bibitem[{Sutton et~al.(2012)Sutton, McCallum et~al.}]{sutton2012introduction}
Charles Sutton, Andrew McCallum, et~al. 2012.
\newblock An introduction to conditional random fields.
\newblock \emph{Foundations and Trends{\textregistered} in Machine Learning},
  4(4):267--373.

\bibitem[{Vaswani et~al.(2017)Vaswani, Shazeer, Parmar, Uszkoreit, Jones,
  Gomez, Kaiser, and Polosukhin}]{vaswani2017attention}
Ashish Vaswani, Noam Shazeer, Niki Parmar, Jakob Uszkoreit, Llion Jones,
  Aidan~N Gomez, {\L}ukasz Kaiser, and Illia Polosukhin. 2017.
\newblock Attention is all you need.
\newblock In \emph{Advances in neural information processing systems}, pages
  5998--6008.

\bibitem[{Vinyals et~al.(2016)Vinyals, Blundell, Lillicrap, Wierstra
  et~al.}]{vinyals2016matching}
Oriol Vinyals, Charles Blundell, Timothy Lillicrap, Daan Wierstra, et~al. 2016.
\newblock Matching networks for one shot learning.
\newblock In \emph{Advances in neural information processing systems}, pages
  3630--3638.

\bibitem[{Wang et~al.(2005)Wang, Deng, and Acero}]{wang2005spoken}
Ye-Yi Wang, Li~Deng, and Alex Acero. 2005.
\newblock Spoken language understanding--an introduction to the statistical
  framework.
\newblock \emph{IEEE Signal Processing Magazine}, 22(5):16--31.

\bibitem[{Yan et~al.(2018)Yan, Zheng, and Cao}]{yan2018few}
Leiming Yan, Yuhui Zheng, and Jie Cao. 2018.
\newblock Few-shot learning for short text classification.
\newblock \emph{Multimedia Tools and Applications}, 77(22):29799--29810.

\bibitem[{Yao et~al.(2014)Yao, Peng, Zweig, Yu, Li, and Gao}]{yao2014recurrent}
Kaisheng Yao, Baolin Peng, Geoffrey Zweig, Dong Yu, Xiaolong Li, and Feng Gao.
  2014.
\newblock Recurrent conditional random field for language understanding.
\newblock In \emph{2014 IEEE International Conference on Acoustics, Speech and
  Signal Processing (ICASSP)}, pages 4077--4081.

\bibitem[{Yoon et~al.(2019)Yoon, Seo, and Moon}]{yoon2019tapnet}
Sung~Whan Yoon, Jun Seo, and Jaekyun Moon. 2019.
\newblock {TapNet}: Neural network augmented with task-adaptive projection for
  few-shot learning.
\newblock \emph{arXiv preprint arXiv:1905.06549}.

\bibitem[{Yu et~al.(2018)Yu, Guo, Yi, Chang, Potdar, Cheng, Tesauro, Wang, and
  Zhou}]{yu2018diverse}
Mo~Yu, Xiaoxiao Guo, Jinfeng Yi, Shiyu Chang, Saloni Potdar, Yu~Cheng, Gerald
  Tesauro, Haoyu Wang, and Bowen Zhou. 2018.
\newblock Diverse few-shot text classification with multiple metrics.
\newblock \emph{arXiv preprint arXiv:1805.07513}.

\bibitem[{Zeldes(2017)}]{zeldes2017gum}
Amir Zeldes. 2017.
\newblock The {GUM} corpus: creating multilayer resources in the classroom.
\newblock \emph{Language Resources and Evaluation}, 51(3):581--612.

\end{thebibliography}
\bibliographystyle{acl_natbib}

\appendix

\section{Detail of Dataset}
\label{app:sec:data_statistic}

The data split method provided by \citet{hou2020few} are applied in SNIPS and NER datasets. Statistical analyses of the original datasets are provided in Table \ref{tab:ori_dataset}, where the number of labels (``\# Labels'') is counted in inside/outside/beginning (IOB) schema. 

 \begin{table}[htbp]
    \small
    \centering
\begin{tabular}{ccccc}
    \hline
    \textbf{Task} & \textbf{Dataset} & \textbf{Domain} & \textbf{\# Sent} & \textbf{\# Labels} \\
    \hline
    \multirow{7}{*}{\begin{tabular}[c]{@{}c@{}}Slot\\ Tagging\end{tabular}} & \multirow{7}{*}{SNIPS} & We & 2100 & 17 \\
    &  & Mu & 2100 & 18 \\
    &  & Pl & 2042 & 10 \\
    &  & Bo & 2056 & 12 \\
    &  & Se & 2059 & 15 \\
    &  & Re & 2073 & 28 \\
    &  & Cr & 2054 & 5 \\
    \hline
    \multirow{4}{*}{NER} & CoNLL & News & 20679 & 9 \\
     & GUM & Wiki & 3493 & 23 \\
     & WNUT & Social & 5657 & 13 \\
     & OntoNotes & Mixed & 159615 & 37 \\
    \hline
\end{tabular}
    \caption{Statistics of original dataset.} 
    \label{tab:ori_dataset}
\end{table}

\citet{hou2020few} reorganized the dataset for few-shot slot tagging and NER in the \emph{episode} data setting~\cite{vinyals2016matching}, where each episode contains a support set (1-shot or 5-shot) and a batch of labeled samples. The 1-shot and 5-shot scenarios mean each label of a domain appears about 1 and 5 times, respectively. The overview of the few-shot data split on SNIPS and NER are shown in Table \ref{tab:few_shot_SNIPS_dataset} and Table \ref{tab:few_shot_NER_dataset} respectively. For SNIPS, each domain consists of 100 episodes. For NER, each domain contains 200 episodes in 1-shot scenario and 100 episodes in 5-shot scenario.

\begin{table}[htbp]
    \small
    \centering
\begin{tabular}{ccccc}
    \hline
    \multirow{2}{*}{\textbf{Domain}} & \multicolumn{2}{c}{\textbf{1-shot}} & \multicolumn{2}{c}{\textbf{5-shot}} \\
    \cline{2-3} \cline{4-5}
    & \textbf{Avg. $|S|$} & \textbf{\# Sent} & \textbf{Avg. $|S|$} & \textbf{\# Sent} \\
    \hline
    \textbf{We} & 6.15 & 2000 & 28.91 & 1000 \\
    \textbf{Mu} & 7.66 & 2000 & 34.43 & 1000 \\
    \textbf{Pl} & 2.96 & 2000 & 13.84 & 1000 \\
    \textbf{Bo} & 4.34 & 2000 & 19.83 & 1000 \\
    \textbf{Se} & 4.29 & 2000 & 19.27 & 1000 \\
    \textbf{Re} & 9.41 & 2000 & 41.58 & 1000 \\
    \textbf{Cr} & 1.30 & 2000 & 5.28 & 1000 \\
    \hline
\end{tabular}
    \caption{Overview of few-shot slot tagging data from SNIPS. ``Avg. $|S|$'' refers to the average support set size of each domain, and ``Sample'' indicates the number of labelled samples in the batches of all episodes.} 
    \label{tab:few_shot_SNIPS_dataset}
\end{table}

\begin{table}[htbp]
    \small
    \centering
\begin{tabular}{ccccc}
    \hline
    \multirow{2}{*}{\textbf{Domain}} & \multicolumn{2}{c}{\textbf{1-shot}} & \multicolumn{2}{c}{\textbf{5-shot}} \\
    \cline{2-3} \cline{4-5}
    & \textbf{Avg. $|S|$} & \textbf{\# Sent} & \textbf{Avg. $|S|$} & \textbf{\# Sent} \\
    \hline
    \textbf{News} & 3.38 & 4000 & 15.58 & 1000 \\
    \textbf{Wiki} & 6.50 & 4000 & 27.81 & 1000 \\
    \textbf{Social} & 5.48 & 4000 & 28.66 & 1000 \\
    \textbf{Mixed} & 14.38 & 4000 & 62.28 & 1000 \\
    \hline
\end{tabular}
    \caption{Overview of few-shot data for NER experiments.} 
    \label{tab:few_shot_NER_dataset}
\end{table}

\section{Training Details}
\label{app:sec:training}

In all the experiments, we use the uncased \texttt{BERT-Base}~\cite{devlin2019bert} as $E$ to extract contextual word embeddings. The models are trained using ADAM \cite{kingma2014adam} with the learning rate of 1e-5 and updated after each episode. We fine-tune BERT with layer-wise learning rate decay (rate is 0.9), i.e., the parameters of the $l$-th layer get an adaptive learning rate $1\text{e-}5* 0.9^{(L - l)}$, where $L$ is the total number of layers in BERT. For the CRF transition parameters, they are initialized as zeros, and a large learning rate of 1e-3 is applied.

For each dataset, we follow \citet{hou2020few} to select one target domain for evaluation, one domain for validation, and utilize the rest domains as source domains for training. The models are trained for five iterations, and we save the parameters with the best F$_1$ score on the validation domain. We use the average F$_1$ score at episode level, and the F$_1$-score is calculated using CoNLL evaluation script\footnote{\url{https://www.clips.uantwerpen.be/conll2000/chunking/output.html}}. For each experiment, we run it ten times with different random seeds generated at \url{https://www.random.org}. 

We run our models on GeForce GTX 2080 Ti Graphics Cards, and the average training time for each epoch and number of parameters in each model are provided in Table \ref{tab:reproducibility}.

\begin{table}[htp]
    \small
    \centering
    \begin{tabular}{|c|c|c|c|}
        \hline
        \multirow{2}{*}{\textbf{Method}} & \multicolumn{2}{c|}{\textbf{Time per Batch}} & \multirow{2}{*}{\textbf{\# Param.}} \\
        \cline{2-3}
        & SNIPS & NER & \\
        \hline
        L-TapNet+CDT+VP & 224ms & 273ms & 110M \\
        ProtoNet+CDT+VP & 176ms & 223ms & 110M \\
        ProtoNet+CDT+VPB & 184ms & 240ms & 110M \\
        \hline    
    \end{tabular}
    \caption{Runtime and mode size of our methods.}
    \label{tab:reproducibility}
\end{table}

\section{Additional Analyses and Results}

\subsection{Fine-tuning on the Support Set}
\label{app:sec:finetune}

Almost all few-shot slot tagging methods choose not to keep fine-tuning on the support set for efficiencies. Here we want to know how performances change if our methods are fine-tuned on the support set. Concretely, pre-trained models are fine-tuned on the support set of one episode and then evaluated on the data batch of the episode. Since different episodes are independent, models would be reinitialized as the pre-trained ones to prepare for the next episode. We fine-tune the ``ProtoNet+CDT+VP'' model for $1\sim10$ steps using the same hyper-parameters with the training. As illustrated in Table \ref{tab:finetune}, we can find that fine-tuning on the support set can get further improvements greatly.

 \begin{table}[htbp]
    \small
    \centering
\begin{tabular}{ccccc}
    \hline
    \multirow{2}{*}{\textbf{Fine-tune step}} & \multicolumn{2}{c}{\textbf{SNIPS}} & \multicolumn{2}{c}{\textbf{NER}} \\
    \cline{2-3} \cline{4-5}
    & \textbf{1-shot} & \textbf{5-shot} & \textbf{1-shot} & \textbf{5-shot} \\
    \hline
    0 & 72.37 & 79.37 & 28.25 & 32.98 \\
    1 & 73.47 & 80.91 & 29.16 & 34.77 \\
    3 & 74.92 & 82.98 & 30.76 & 37.49 \\
    5 &  75.48 & 83.97 & 31.93 & 39.29 \\
    10 &  75.72 & 84.87 & 33.41 & 42.03 \\
    \hline
\end{tabular}
    \caption{Results are averaged F1-scores of all domains. The backbone method is ``ProtoNet+CDT+VP''.} 
    \label{tab:finetune}
\end{table}

\begin{table*}[t]
    \scriptsize
    \centering
\begin{tabular}{ccccccccc}
    \hline
    \textbf{Model} & \textbf{We} & \textbf{Mu} & \textbf{Pl} & \textbf{Bo} & \textbf{Se} & \textbf{Re} & \textbf{Cr} & \textbf{Avg.} \\
    \hline
    SimBERT$^*$ & 36.10$\pm$0.00 & 37.08$\pm$0.00 & 35.11$\pm$0.00 & 68.09$\pm$0.00 & 41.61$\pm$0.00 & 42.82$\pm$0.00 & 23.91$\pm$0.00 & 40.67$\pm$0.00 \\
    TransferBERT$^*$ & 55.82$\pm$2.75 & 38.01$\pm$1.74 & 45.65$\pm$2.02 & 31.63$\pm$5.32 & 21.96$\pm$3.98 & 41.79$\pm$3.81 & 38.53$\pm$7.42 & 39.06$\pm$3.86 \\
    L-WPZ(ProtoNet)+CDT+PWE$^*$ & 71.23$\pm$6.00	& 47.38$\pm$4.18	& 59.57$\pm$5.55	& 81.98$\pm$2.08	& 69.83$\pm$1.94	& 66.52$\pm$2.72	& 62.84$\pm$0.58	& 65.62$\pm$3.29 \\
    L-TapNet+CDT+PWE$^*$ & 71.53$\pm$4.04	& {60.56}$\pm$0.77	& 66.27$\pm$2.71	& \textbf{84.54}$\pm$1.08	& 76.27$\pm$1.72	& 70.79$\pm$1.60	& 62.89$\pm$1.88	& 70.41$\pm$1.97 \\
    \hline
    L-TapNet+CDT+VP & 71.65$\pm$1.30 & \textbf{61.73}$\pm$1.49 & 63.97$\pm$0.84 & 83.34$\pm$0.65 & 74.00$\pm$1.01 & 71.91$\pm$0.97 & 71.02$\pm$1.47 & 71.09$\pm$1.10  \\
    ProtoNet+CDT+VP & \textbf{73.56}$\pm$0.93 & 58.40$\pm$1.11 & {68.93}$\pm$0.95 & 82.32$\pm$0.78 & \textbf{79.69}$\pm$0.55 & \textbf{73.40}$\pm$0.75 & \textbf{70.25}$\pm$1.22 & \textbf{72.37}$\pm$0.90 \\
    L-ProtoNet+CDT+VP & 73.19$\pm$1.65 & 58.62$\pm$1.02 & 68.26$\pm$0.42 & 83.54$\pm$0.62 & 77.88$\pm$0.59 & 73.48$\pm$1.13 & 69.54$\pm$1.64 & 72.07$\pm$1.01 \\
    ProtoNet+CDT+VPB & 72.65$\pm$1.30 & 57.35$\pm$0.59 & 68.72$\pm$0.52 & 81.92$\pm$0.72 & 74.68$\pm$0.54 & 72.48$\pm$0.94 & 70.04$\pm$2.05 & 71.12$\pm$0.95 \\
    L-ProtoNet+CDT+VPB & 73.12$\pm$1.30 &  57.86$\pm$0.53 & \textbf{69.01}$\pm$0.35 & 82.49$\pm$0.68  & 75.11$\pm$0.70 & 73.34$\pm$0.89 &  70.46$\pm$1.22 & 71.63$\pm$0.81 \\
    \hline
\end{tabular}
    \caption{F$_1$ scores on 1-shot slot tagging of SNIPS dataset. * indicates a result borrowed from \citet{hou2020few}.} 
    \label{tab:1shot_snips_with_var}
\end{table*}

\begin{table*}[t]
    \scriptsize
    \centering
\begin{tabular}{ccccccccc}
    \hline
    \textbf{Model} & \textbf{We} & \textbf{Mu} & \textbf{Pl} & \textbf{Bo} & \textbf{Se} & \textbf{Re} & \textbf{Cr} & \textbf{Avg.} \\ 
    \hline
    SimBERT$^*$ & 53.46$\pm$0.00 & 54.13$\pm$0.00 & 42.81$\pm$0.00 & 75.54$\pm$0.00 & 57.10$\pm$0.00 & 55.30$\pm$0.00 & 32.38$\pm$0.00 & 52.96$\pm$0.00 \\
    TransferBERT$^*$ & 59.41$\pm$0.30 & 42.00$\pm$2.83 & 46.07$\pm$4.32 & 20.74$\pm$3.36 & 28.20$\pm$0.29 & 67.75$\pm$1.28 & 58.61$\pm$3.67 & 46.11$\pm$2.29 \\
    L-WPZ(ProtoNet)+CDT+PWE$^*$ & 74.68$\pm$2.43 & 56.73$\pm$3.23 & 52.20$\pm$3.22 & 78.79$\pm$2.11 & 80.61$\pm$2.27 & 69.59$\pm$2.78 & 67.46$\pm$1.91 & 68.58$\pm$2.56 \\
    L-TapNet+CDT+PWE$^*$ & 71.64$\pm$3.62 & 67.16$\pm$2.97 & 75.88$\pm$1.51 & 84.38$\pm$2.81 & 82.58$\pm$2.12 & 70.05$\pm$1.61 & 73.41$\pm$2.61 & 75.01$\pm$2.46 \\
    \hline
    L-TapNet+CDT+VP & 78.25$\pm$1.31 & {67.79}$\pm$1.18 & 70.66$\pm$2.11 & 86.17$\pm$1.16 & 75.80$\pm$1.61 & 78.51$\pm$1.28 & 75.93$\pm$1.20 & 76.16$\pm$1.41  \\
    ProtoNet+CDT+VP & {79.88}$\pm$0.76 & {67.77}$\pm$0.73 & {78.08}$\pm$1.28 & {87.68}$\pm$0.40 & \textbf{86.59}$\pm$0.68 & {79.95}$\pm$0.45 & {75.61}$\pm$1.88 & {79.37}$\pm$0.88 \\
    L-ProtoNet+CDT+VP & 80.26$\pm$0.78 & 67.81$\pm$0.59 & 74.62$\pm$1.37 & 88.16$\pm$0.48 & 85.79$\pm$0.71 & 80.41$\pm$0.65 & 73.84$\pm$1.68 & 78.70$\pm$0.89 \\
    ProtoNet+CDT+VPB & {82.91}$\pm$0.85 & {69.23}$\pm$0.56 & {80.85}$\pm$1.18 & {90.69}$\pm$0.43 & 86.38$\pm$0.47 & {81.20}$\pm$0.45 & \textbf{76.75}$\pm$1.59 & {81.14}$\pm$0.79 \\
    L-ProtoNet+CDT+VPB & \textbf{82.93}$\pm$0.59 & \textbf{69.62}$\pm$0.46 & \textbf{80.86}$\pm$1.04 & \textbf{91.19}$\pm$0.37 & 86.58$\pm$0.63 & \textbf{81.97}$\pm$0.57 & 76.02$\pm$1.65 & \textbf{81.31}$\pm$0.76 \\
    \hline
\end{tabular}
    \caption{F$_1$ scores on 5-shot slot tagging of SNIPS dataset. * indicates a result borrowed from \citet{hou2020few}.} 
    \label{tab:5shot_snips_with_var}
\end{table*}

\begin{table*}[]
    \small
    \centering
\begin{tabular}{cccccc}
    \hline
    {\textbf{Model}} & \textbf{News} & \textbf{Wiki} & \textbf{Social} & \textbf{Mixed} & \textbf{Avg.} \\
    \hline
    SimBERT$^*$ & 19.22$\pm$0.00 & 6.91$\pm$0.00 & 5.18$\pm$0.00 & 13.99$\pm$0.00 & 11.32$\pm$0.00 \\
    TransferBERT$^*$ & 4.75$\pm$1.42 & 0.57$\pm$0.32 & 2.71$\pm$0.72 & 3.46$\pm$0.54 & 2.87$\pm$0.75 \\
    L-TapNet+CDT+PWE$^*$ & 44.30$\pm$3.15 & \textbf{12.04}$\pm$0.65 & 20.80$\pm$1.06 & 15.17$\pm$1.25 & 23.08$\pm$1.53 \\
    \hline
    L-TapNet+CDT+VP & 44.73$\pm$2.56 & 8.91$\pm$0.58  & \textbf{30.61}$\pm$0.66 & {29.39}$\pm$1.26 & {28.41}$\pm$1.26  \\
    ProtoNet+CDT+VP & {44.82}$\pm$1.62 & 11.32$\pm$0.29 & {26.96}$\pm$0.54 & {29.91}$\pm$1.23 & {28.25}$\pm$0.92  \\
    L-ProtoNet+CDT+VP & \textbf{45.93}$\pm$1.90 & 8.76$\pm$0.18  & 29.21$\pm$1.06 & 32.44$\pm$1.19 & \textbf{29.09}$\pm$1.08  \\
    ProtoNet+CDT+VPB & 42.50$\pm$0.72 & 10.78$\pm$0.32 & 27.17$\pm$0.66 & {32.06}$\pm$1.89 & 28.13$\pm$0.90 \\
    L-ProtoNet+CDT+VPB & 43.47$\pm$0.58 & 10.95$\pm$0.28 & 28.43$\pm$0.45 & \textbf{33.14}$\pm$1.88 & 29.00$\pm$0.80 \\
    \hline
\end{tabular}
    \caption{F$_1$ scores on 1-shot slot tagging of NER dataset. * indicates a result borrowed from \citet{hou2020few}.} 
    \label{tab:1shot_ner_with_var}
\end{table*}

\begin{table*}[]
    \small
    \centering
\begin{tabular}{cccccc}
    \hline
    \textbf{Model} & \textbf{News} & \textbf{Wiki} & \textbf{Social} & \textbf{Mixed} & \textbf{Avg.} \\
    \hline
    SimBERT$^*$ & 32.01$\pm$0.00 & 10.63$\pm$0.00 & 8.20$\pm$0.00 & 21.14$\pm$0.00 & 18.00$\pm$0.00 \\
    TransferBERT$^*$ & 15.36$\pm$2.81 & 3.62$\pm$0.57 & 11.08$\pm$0.57 & {35.49}$\pm$7.60 & 16.39$\pm$2.89 \\
    L-TapNet+CDT+PWE$^*$ & 45.35$\pm$2.67 & 11.65$\pm$2.34 & 23.30$\pm$2.80 & 20.95$\pm$2.81 & 25.31$\pm$2.65 \\
    \hline
    L-TapNet+CDT+VP & 50.43$\pm$1.62 & 8.41$\pm$0.53  & {29.93}$\pm$1.12 & {37.59}$\pm$1.98 & 31.59$\pm$1.31   \\
    ProtoNet+CDT+VP & {54.82}$\pm$0.53 & {16.30}$\pm$0.55 & {27.43}$\pm$0.51 & 33.38$\pm$0.76 & {32.98}$\pm$0.59 \\
    L-ProtoNet+CDT+VP & 55.68$\pm$0.84 & 10.39$\pm$0.23 & 31.39$\pm$0.85 & 37.83$\pm$1.50 & 33.82$\pm$0.86 \\
    ProtoNet+CDT+VPB & \textbf{57.42}$\pm$1.36 & \textbf{19.48}$\pm$0.28 & {35.06}$\pm$0.63 & {44.45}$\pm$1.01 & \textbf{39.10}$\pm$0.82 \\
    L-ProtoNet+CDT+VPB & 56.30$\pm$1.76 & 18.57$\pm$0.49 & \textbf{35.42}$\pm$0.47 & \textbf{44.71}$\pm$0.92 & 38.75$\pm$0.91 \\
    \hline
\end{tabular}
    \caption{F$_1$ scores on 5-shot slot tagging of NER dataset. * indicates a result borrowed from \citet{hou2020few}.} 
    \label{tab:5shot_ner_with_var}
\end{table*}

\subsection{Result with Standard Deviations}
\label{app:sec:res_std}

Table \ref{tab:1shot_snips_with_var}, \ref{tab:5shot_snips_with_var}, \ref{tab:1shot_ner_with_var} and \ref{tab:5shot_ner_with_var} show the complete results with standard deviations on SNIPS and NER.

\end{document}